\pdfoutput=1
\documentclass[10pt,twocolumn,letterpaper]{article}

\usepackage{cvpr}              % To produce the CAMERA-READY version
\makeatletter
\renewcommand\paragraph{%
  \@startsection{paragraph}{4}{\z@}%
    {3ex \@plus 0.5ex \@minus .2ex}%    % Before-skip
    {-0.5em}%                           % After-skip (negative = no line break)
    {\normalfont\normalsize\bfseries}%
}
\makeatother

\usepackage{multirow}

\definecolor{cvprblue}{rgb}{0.21,0.49,0.74}
\usepackage[pagebackref,breaklinks,colorlinks,allcolors=cvprblue]{hyperref}
\usepackage{setspace}
\def\paperID{9854} % *** Enter the Paper ID here
\def\confName{CVPR}
\def\confYear{2026}

\title{ViHOI: Human-Object Interaction Synthesis with Visual Priors}

\author{
    Songjin Cai \quad Linjie Zhong \quad Ling Guo \quad  Changxing Ding\textsuperscript{*} \\
    South China University of Technology \\
    {\tt\small \{eecaisongjin, eelinjie, eeguoling\}@mail.scut.edu.cn, chxding@scut.edu.cn}
}

\begin{document}

\captionsetup[figure]{hypcap=false}

\newcommand{\teasercaption}{
We propose ViHOI, a novel plug‑and‑play approach that enables motion diffusion models to effectively leverage rich visual priors from a set of 2D reference images. It utilizes reference images synthesized by a text‑to‑image generation model that contains rich world knowledge during inference, enabling strong generalization to unseen objects and delivering superior results across multiple benchmarks.}

% \vspace{-4mm}
% \twocolumn[{
%     \renewcommand\twocolumn[1][]{#1}
%     \maketitle
%     \centering

%     \begin{minipage}{1.0\textwidth}
%         \centering
%         \includegraphics[width=1.0\textwidth,height=0.415\textheight]{figures/cover.pdf}
%     \end{minipage}

%     \captionof{figure}{\teasercaption}
%     \vspace{4mm}
%     \label{fig:firstfigure}
% }
% ]
%     %
\twocolumn[{
\renewcommand\twocolumn[1][]{#1}
\maketitle
\vspace{-9mm}
\begin{center}
    \centering
    \captionsetup{type=figure}
    \includegraphics[width=1\linewidth]{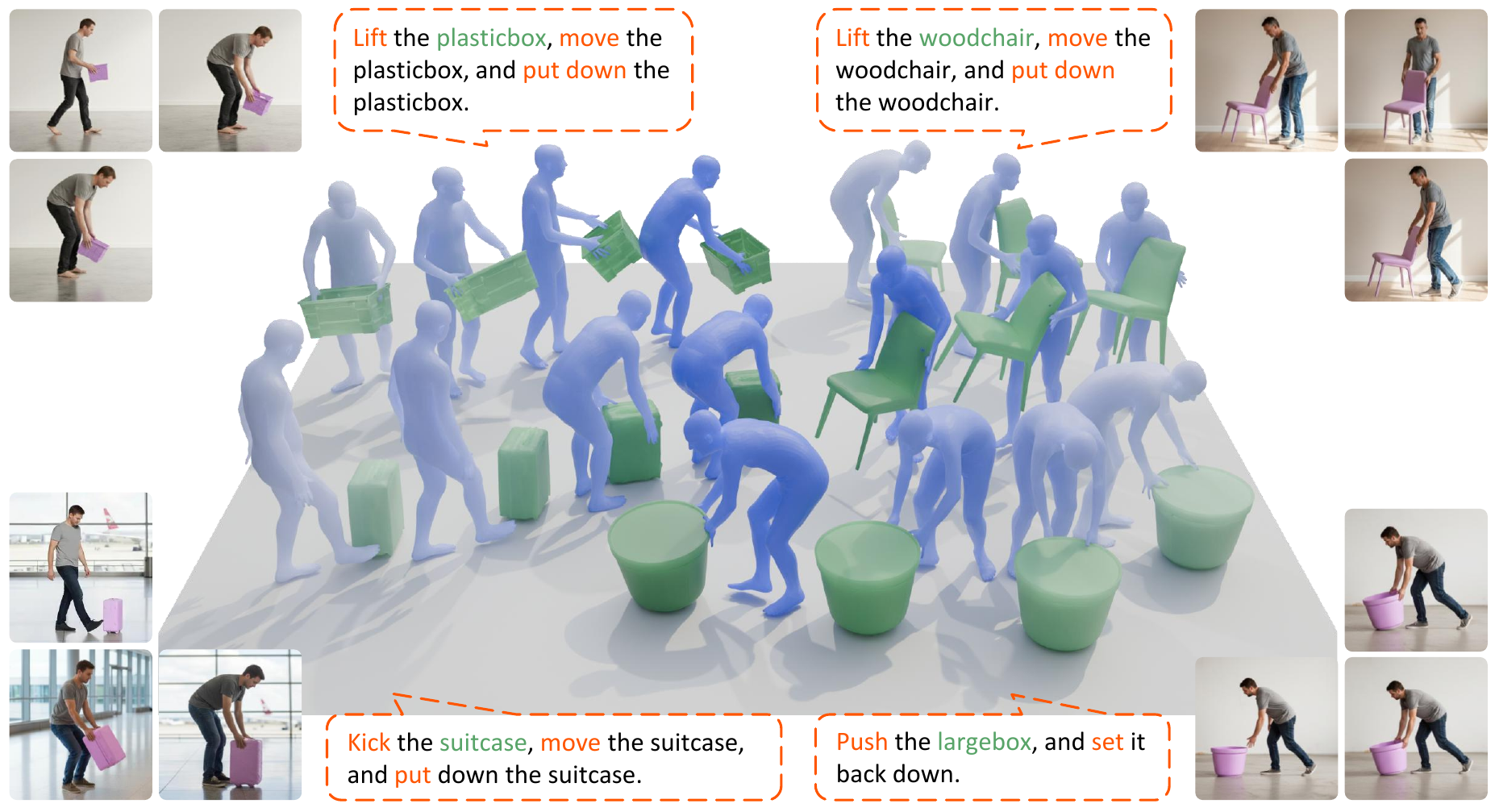}
    \vspace{-6mm}

    \captionof{figure}{We propose ViHOI, a novel plug‑and‑play approach that enables motion diffusion models to effectively leverage rich visual priors from a set of 2D reference images. It utilizes reference images synthesized by a text‑to‑image generation model that contains rich world knowledge during inference, enabling strong generalization to unseen objects and delivering superior results across multiple benchmarks.}
    \label{fig:firstfigure}
    \vspace{1mm}
\end{center}
}]

\maketitle
\renewcommand{\thefootnote}{\fnsymbol{footnote}} 
\footnotetext[1]{Corresponding author.} 
\begin{abstract}
\quad Generating realistic and physically plausible 3D Human-Object Interactions (HOI) remains a key challenge in motion generation. One primary reason is that describing these physical constraints with words alone is difficult. To address this limitation, we propose a new paradigm: extracting rich interaction priors from easily accessible 2D images. Specifically, we introduce ViHOI, a novel framework that enables diffusion-based generative models to leverage rich, task-specific priors from 2D images to enhance generation quality. We utilize a large Vision-Language Model (VLM) as a powerful prior-extraction engine and adopt a layer-decoupled strategy to obtain visual and textual priors. Concurrently, we design a Q-Former-based adapter that compresses the VLM’s high-dimensional features into compact prior tokens, which significantly facilitates the conditional training of our diffusion model. Our framework is trained on motion-rendered images from the dataset to ensure strict semantic alignment between visual inputs and motion sequences. During inference, it leverages reference images synthesized by a text-to-image generation model to improve generalization to unseen objects and interaction categories. Experimental results demonstrate that ViHOI achieves state-of-the-art performance, outperforming existing methods across multiple benchmarks and demonstrating superior generalization. The code for this work will be released at~\href{https://github.com/MPI-Lab/ViHOI}{https://github.com/MPI-Lab/ViHOI}. 
\end{abstract}
    
\section{Introduction}
\label{sec:intro}

Human-Object Interaction (HOI) generation aims to synthesize realistic, physically plausible, and semantically consistent interaction sequences between humans and objects \cite{HOI_appplication2, AvatarGo_HOI_appplication4}. It holds significant promise for applications in virtual reality\cite{VR}, computer animation \cite{HOI_appplication1,HOI_appplication2, HOI_appplication3}, and robotics \cite{Robot_HOI_appplication5,MotionGPT3}. However, generating high-quality HOI sequences remains a considerable challenge. These applications demand that the generated motion not only adheres strictly to the textual prompts but also satisfies rigorous physical constraints \cite{Full-Body_Articulated_Interaction,HOI_text}. 

With the rapid development of diffusion models, there has been an increasing effort to apply them to the HOI generation task \cite{Diffusion, 2025_Model_object_affordance_w_VeidoDiffusion}. However, their performance is mainly limited by the quality of the conditioning signals they received \cite{Diffusion_Sora,Diffusion_5}. The HOI process involves a continuous series of spatial state changes and must adhere to reasonable interaction relationships between the human and the object \cite {HOIAnimator, ARCTIC,Interdiff}. However, the text annotations in existing datasets typically provide only an abstract description of the HOI sequence \cite{SemGeoMo,F-HOI}. Although intuitive, they lack crucial geometric and spatial priors. For example, ``pick up a box" provides no specific details regarding the box's shape, size, or the required human pose. This forces the model into a complex ``one-to-many" learning problem, which not only hinders the realism and controllability of the generated motion but also weakens the model's ability to generalize to objects and actions outside the dataset \cite{TriDiHOI,NIFTY}.

To overcome this challenge, existing works incorporate various priors into the generation process. These approaches generally fall into two paradigms: (1) semantic enhancement, which leverages large language models (LLMs) to enrich the text annotations \cite{SemGeoMo, F-HOI}, and (2) physical constraint, which introduces explicit priors based on geometry and interaction \cite{ChainHOI, CG_HOI, HOI-Diff, HOI-Dyn}. For semantic enhancement, fine-grained text improves motion quality, but still lacks structured knowledge to precisely coupling motions with the geometry of objects, leading to suboptimal performance on unseen objects and motions \cite{HUMANISE}.
In the case of physical constraints, some works introduce explicit contact priors, such as affordance maps or contact points \cite{CG_HOI}. However, these methods tend to focus on the immediate interaction regions, neglecting the global dynamics and coherence of the full-body motion \cite{HOI-Diff}.
We argue that these approaches overlook a robust and readily accessible source of information: 2D images. We believe that 2D images provide a rich set of visual interaction priors, such as object shape, scale, and human-object spatial relations. Leveraging these visual priors may significantly enhance the fidelity and physical plausibility of the HOI generation model.

Accordingly, we propose \textbf{ViHOI}, a novel framework that enables diffusion-based generative models to leverage these rich, task-specific priors from 2D images to enhance generation quality. Our framework consists of two core components: \textbf{VLM-based Prior Extractor} and \textbf{Vision-aware HOI Generator}. Specifically, we use a large VLM \cite{DBLP:journals/arXiv/abs-2502-13923} and design a structured prompt to explicitly guide it to focus on key human-object interaction cues in the images, such as the object's shape, size, and human-object interaction pose.
We simultaneously extract both visual and textual information from the VLM. This approach inherently ensures the semantic alignment of these two distinct modalities. Furthermore, we observed that different layers of the VLM exhibit varying levels of attention to images versus text \cite{LED,OPENHELIX,InteractVLM5}. Therefore, we adopt a layer-decoupled strategy to extract the VLM's outputs.
Specifically, we extract the VLM's intermediate outputs: (1) an embedding from an early, vision-friendly layer serves as the spatial-visual prior, capturing rich geometric detail; and (2) an embedding from a deeper, semantic-rich layer serves as the semantic control, which is extracted from the tokens corresponding to the text description within our prompt.

For the Vision-aware HOI Generator, we use a diffusion-based motion generator that injects visual and textual priors via self-attention. 
To mitigate the impact of redundant features from the VLM's intermediate layers, we design two Q-Former-based prior adapters \cite{BLIP-Q-Former}. These adapters refine high-dimensional visual and textual embeddings into structured prior tokens, making them suitable for downstream generation tasks. 
Our framework employs different strategies to obtain visual priors for training and inference. During training, we extract visual priors by rendering 2D images directly from the Ground Truth (GT) motion sequences in our dataset. To capture the interaction's dynamics, we utilize the dataset's contact labels to select three keyframes corresponding to the start, middle, and end frames of the contact phase. This strategy ensures strict semantic consistency between the visual prior and the target motion at a low cost.
During inference, we leverage an advanced text-to-image generation model to synthesize three sequentially coherent and reasonable HOI images. This leverages the rich world knowledge embedded in the image generation model, thereby enhancing our motion generation model's generalization capability to unseen objects and scenarios.

Despite the inherent style gap between the clean, rendered training images and the synthesized inference images, our model maintains robust performance and strong generalization during testing. This demonstrates that our VLM-based prior extractor successfully identifies and leverages the underlying motion-relevant features, proving the robustness and effectiveness of our proposed image-as-motion-prior paradigm.

We further showcase the versatility of our method as a plug-and-play module, successfully boosting the performance of various off-the-shelf HOI motion synthesis models \cite{CHOIS,ROG,MDM}. Finally, our method achieves state-of-the-art performance in both qualitative and quantitative terms, advancing the field of HOI motion synthesis.

\section{Related Work}
To compensate for the limitations of text-only conditions, which omit key details like shape, size, and contact, a significant body of work has sought to inject additional prior knowledge into the generation process. These approaches generally fall into two main paradigms: (1) semantic enhancement \cite{F-HOI, SemGeoMo, InterAct}, and (2) physical constraints \cite{CG_HOI, ChainHOI, SemGeoMo, HOI-Dyn, HOI-Diff, HOI-RMD}. The semantic enhancement approach addresses the problem within the language modality, leveraging LLM to expand simple instructions into detailed scripts with explicit steps and descriptions, providing more fine-grained textual guidance for generation. However, due to the lack of direct supervision over the spatial relationship between the human and the object, the resulting generations may suffer from inconsistent contact, interpenetration, and floating objects \cite{Gen_HOI,HUMANISE}.

Physical constraints introduce more explicit, non-textual priors to govern the physical interaction process. Current methods include: (i) Contact-based methods, which typically select a fixed set of body joints and use a generative module to predict their contact points on the object \cite{CG_HOI, SemGeoMo}. The contact points often act as a guidance signal during inference \cite{CG_HOI, HOI-Diff}; (ii) Kinematics-based methods, which integrate physical laws into the model, such as kinematic-chain approaches defining joint coordination or physics-driven rules that restrict object motion \cite{ChainHOI, HOI-Dyn, ROG}; and (iii) Decomposition-based methods, which often use LLM planning to decompose a long motion into segments to model human-object relative states. While effective at enforcing specific rules, these priors are usually based on simplified or localized representations \cite{HOI-RMD, MotionGPT3}. Consequently, it may present suboptimal performance in full-body coordinated motion and suffer from a lack of overall coherence.

Recently, an emerging line of work has explored utilizing text-to-video generation models to facilitate human motion synthesis~\cite{T2V_recovery_1,T2V_recovery_2,T2V_recovery_3,T2V_recovery_4}. These approaches typically rely on a complex ``video generation and 3D pose recovery" pipeline. However, such methods are limited by the inherent inaccuracies of 2D-to-3D pose estimation. In particular, lifting 3D poses from generated videos often leads to noticeable jittering and temporal inconsistencies, while also introducing substantial computational overhead

In this paper, we employ robust visual priors extracted from a set of reference images to facilitate HOI motion synthesis. These priors, encoded as compact tokens, implicitly capture key interaction details. Experimental results show that the visual priors yield more realistic generation results and enable our model to achieve superior generalization.

\section{Method}
\begin{figure*}[ht]
  \centering
  \includegraphics[width=\linewidth]{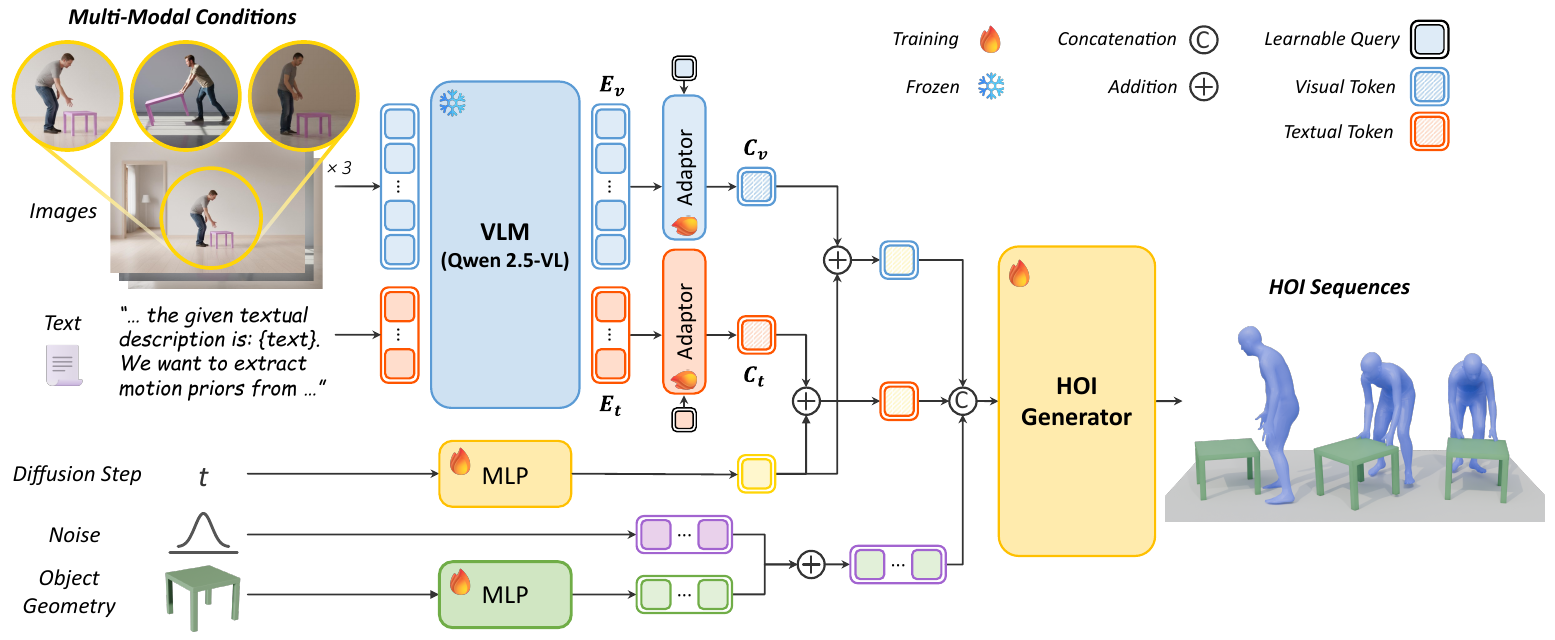}
  \caption{
  Overall architecture of \textbf{ViHOI}. We extract visual priors from a set of reference images and textual priors from the input prompt using a VLM. This allows for natural alignment between priors of the two modalities. Subsequently, two Q-Former-based prior adapters distill these high-dimensional priors into a single compact token, respectively, providing the diffusion model with semantically consistent conditioning signals. At each denoising step, a selected HOI generator uses these compact visual and textual prior tokens to guide the synthesis of realistic, semantically coherent human-object interactions.}
  \label{fig:framework}
\vspace{-3mm}
\end{figure*}
We propose \textbf{ViHOI}, a novel framework that enables diffusion-based generative models to leverage rich, task-specific priors from 2D images to enhance generation quality. Section~\ref{sec:vlm_prior_extractor} details our task-aware prior extractor, which utilizes a VLM to extract decoupled image and text priors. Section~\ref{sec:hoi_generator} describes the vision-aware HOI generator, built upon a Diffusion Transformer (DiT), which effectively fuses these priors to synthesize the final motion. Section~\ref{sec:reference_image} outlines our training and inference strategies, which use rendered images from the ground-truth (GT) motion data during training and leverage a text-to-image synthesis model at inference. An overview of our method is described in Fig.~\ref{fig:framework}.

\subsection{VLM-based Prior Extractor}\label{sec:vlm_prior_extractor}
We use Qwen2.5-VL \cite{DBLP:journals/arXiv/abs-2502-13923} as our prior extractor, leveraging its extensive world knowledge and robust joint vision-language understanding ability to extract visual and textual priors.
Standard vision encoders \cite{CLIP} typically process only a single image at a time, thereby missing the temporal cues that multiple images can provide. In contrast, VLM architecture supports multi-image input, allowing it to capture temporal dynamics and interactive context—thereby supplying downstream generative models with richer, more dynamic visual priors.
Moreover, injecting both visual and textual conditions into a generative model requires that the two modalities be semantically aligned. In our approach, the VLM architecture treats image patches and text tokens as elements of a unified sequence, jointly performing attention across Transformer layers \cite{Qwen2,MLLM}. This co-processing strategy ensures that the two modalities align naturally through the model’s hierarchical reasoning, as substantiated by our ablation studies.
\paragraph{Instruction Design.} 
In addition to human motion, images contain substantial redundancy, such as background, object textures, and the character's clothing. We therefore carefully design a prompt to draw motion cues from the reference images. It explicitly defines the task, indicates the key attributes to focus on (e.g., object shape and contact regions), and incorporates the original text annotation from the dataset. The designed prompt is as follows:

\begin{spacing}{1.1}
 \emph{
``We are conducting the text-to-HOI motion generation task and the given textual description is: \{text\}. We want to extract motion priors from the following three reference images to facilitate the generation of Human-Object-Interaction motion. These priors include the human pose, the object's shape and size, and the contact region on the object during interaction, etc. The initial position of the object is in front of the person."
}
\end{spacing}

Here, \emph{\{text\}} is the original text annotation for one motion sequence in the dataset. We index the start and end tokens of \emph{\{text\}} in the prompt and extract its token embeddings from the VLM as the text-conditioning embeddings. These embeddings are then used together with the image priors to modulate the generator, ensuring that the text instruction and the image prior are aligned.

\vspace{-2mm}
\paragraph{Decoupled Priors Extraction.}We extract both visual and textual priors from the VLM for motion generation. The deeper layers of a VLM enjoy stronger text encoding capabilities, while its shallower layers reserve more visual details \cite{LED}. Based on this insight, we introduce a decoupled strategy to obtain the visual and textual priors. Specifically, we extract visual and textual priors from different layers of the VLM.

\begin{itemize}
    \item \textbf{Visual Prior ($E_v$):} We extract a set of visual embeddings from the 3rd layer of the LLM in Qwen2.5-VL. These embeddings preserve rich geometric and spatial cues from the reference images.
    \item \textbf{Textual Prior ($E_t$):} We employ the embeddings of the text tokens corresponding to \emph{\{text\}} from the 12th layer of the LLM in Qwen2.5-VL. These embeddings represent the textual description of the motion sequence.
\end{itemize}

This decoupled strategy offers the HOI generator more informative priors for each modality. We will compare the performance of different layer combinations in the experimentation section.

\subsection{Vision-aware HOI Generator}\label{sec:hoi_generator}
Our VLM-based prior extractor can be used as a plug-and-play module for the diffusion-based HOI motion generators \cite{MDM, ROG, CHOIS}. The visual and textual priors are transformed into conditional tokens that guide an iterative denoising process for HOI motion synthesis. Specifically, we represent an HOI motion sequence as $x_0 \in \mathbb{R}^{L \times D}$, where $L$ is the sequence length and $D$ is the dimension of the pose representation (including SMPL-X \cite{SMPLX} parameters, object position, and object rotation parameters). Starting with a clean motion sequence $x_0$ sampled from the training data, we apply the diffusion process to generate a sequence of progressively noisier data $\{x_t\}_{t=1}^T$, where $T$ is the total number of diffusion steps. This forward process is defined as:
\begin{equation} \label{eq:1}
q(x_t | x_0) = \mathcal{N}(x_t; \sqrt{\bar{\alpha}_t} x_0, (1 - \bar{\alpha}_t) \mathbf{I}),
\end{equation}
where $\bar{\alpha}_t = \prod_{i=1}^t \alpha_i$, and $\alpha_t \in (0, 1)$ is a time-dependent variance schedule parameter, $\mathcal{N}(\mathbf{0}, \mathbf{I})$ denotes the standard normal distribution, and $\mathbf{I}$ is the identity matrix. As $t$ increases, the distribution of $x_t$ gradually approaches $\mathcal{N}(\mathbf{0}, \mathbf{I})$.

 The HOI motion generation model $f_\theta$ is trained to approximate the reverse diffusion process $p(x_0|c)$, i.e., reconstructing $x_0$ by iteratively denoising from $x_T$ given the condition $c$. The training objective is to minimize the following reconstruction loss:
\begin{equation}\label{eq:2}
    \mathcal{L} = \mathbb{E}_{t, x_0} \left[ \left\| x_0 - f_\theta(x_t, t, c) \right\|^2 \right],
\end{equation}
where $c$ contains both visual and textual conditions from Section~\ref{sec:vlm_prior_extractor}.

\vspace{-2mm}
\paragraph{Prior Adaptors.} The visual and textual priors extracted from the VLM's intermediate layers are high-dimensional, length-variable token sequences. Therefore, it is challenging to use them directly as conditions for the motion diffusion model. To address this problem, we design Q-Former-based prior adaptors to distill these rich priors into compact, fixed-dimensional representations.

Taking $E_v \in \mathbb{R}^{L_v \times d}$ as an example, where
$L_v$ denotes the number of visual tokens, and $d$ presents the token embedding dimension. We first map these token embeddings to the same dimension as the motion tokens in the selected HOI generator:
\begin{equation}\label{eq:3}
Z_v = \text{LayerNorm}(\text{Linear}(E_v)).
\end{equation}
Next, we perform cross-attention between a learnable query ${q_v}$ and ${Z_v}$, where ${Z_v}$ serves as both the key and value:
\begin{equation} \label{eq:4}
    c_v = \text{CrossAttention}(q_v, Z_v, Z_v).
\end{equation}
The Q-Former contains two successive cross-attention layers. Finally, we obtain ${c_v}$, which provides visual priors on interaction from the VLM. Similarly, we obtain a compact textual prior $c_t$ from $E_t$ using another Q-Former with the same structure. Therefore, $c=\{c_v,c_t \}$ in Eq.~\ref{eq:2}.

During training, we freeze the VLM's parameters and jointly train both prior adaptors and the HOI Generator. On the one hand, this strategy preserves the rich world knowledge contained in the VLM. On the other hand, it compels the prior adaptors to extract the most relevant information from the VLM for HOI synthesis.

\subsection{Reference Image Generation}\label{sec:reference_image}
\begin{figure}[ht!]
  \centering
  \includegraphics[width=1.0\linewidth]{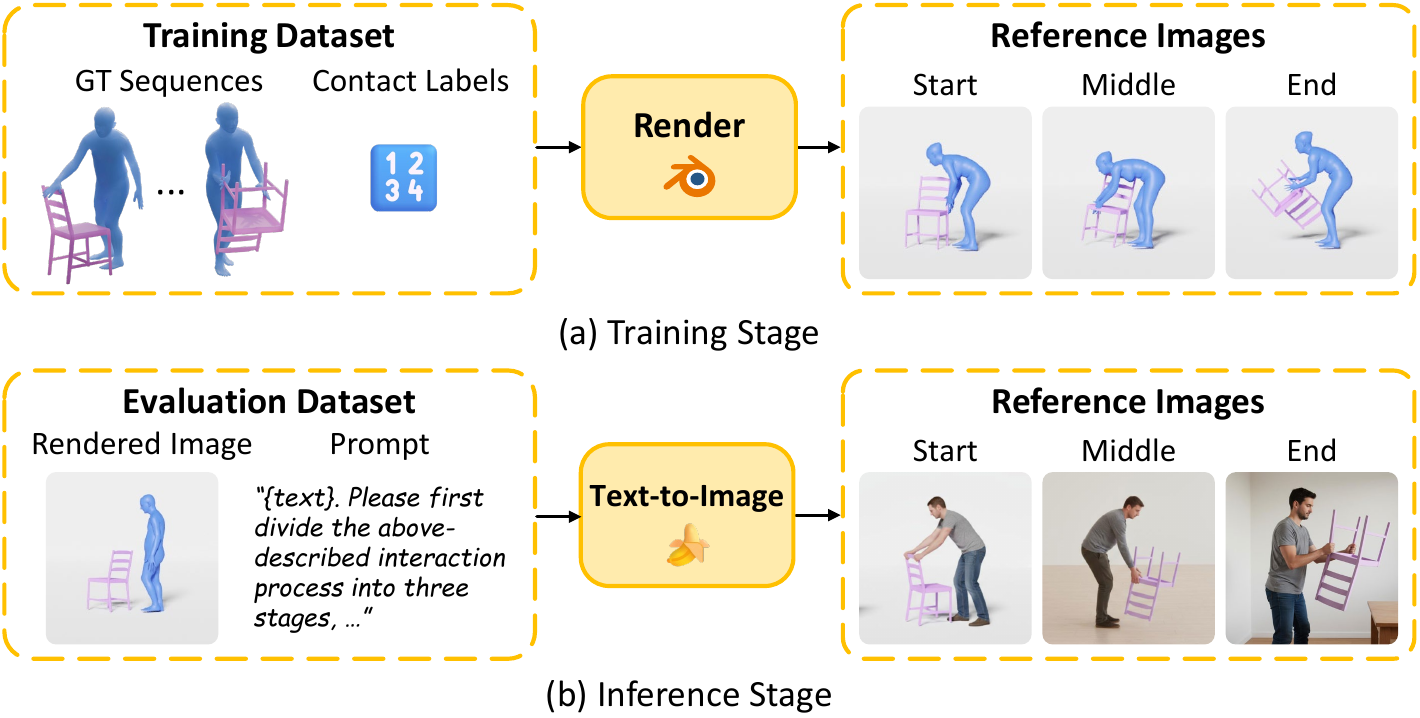}
   \vspace{-6mm}  
  \caption{Illustration of strategies to obtain the set of reference images during 
  the training and inference phases, respectively.
  }
   \vspace{-3mm}
  \label{fig:train_and_inference}
\end{figure}

\textbf{Training Stage.} As illustrated in Fig.~\ref{fig:train_and_inference} (a), we render the ground-truth motion sequences of one dataset to obtain 2D reference images \cite{Blender}. Moreover, we use the contact labels in the training data to select three reference images corresponding to the start, middle, and end frames of the interaction process. This method excels at capturing the interaction’s critical temporal dynamics. More importantly, it enforces strict semantic alignment between the 2D visual condition and the 3D motion sequence, thereby circumventing an extra costly process of curating large-scale image-motion paired data.

\vspace{-3mm}
\paragraph{Inference Stage.} As illustrated in Fig.~\ref{fig:train_and_inference} (b), we use an advanced text-to-image generation model named Nano Banana \cite{Nano_banana} to synthesize three reference images. We provide ``Nano Banana" with a rendered image depicting both the human and the object in a static pose. This enables the model to accurately perceive the object's geometry and the relative scale between the human and the object.
Moreover, we design the following prompt that guides Nano Banana to generate three temporally coherent and semantically correct reference images:

\emph{
``\{text\}. Please first divide the above-described interaction process into three stages, and ensure that there is contact between the character and the object in each stage. Then, synthesize one image for each of the three stages. You should ensure each image contains only one character and one object, and that the object's shape and size match those in the provided image. Moreover, both the background and the character should be realistic and consistent across the three generated images."
}

Similarly, \emph{\{text\}} denotes the original text annotation for the motion sequence in the dataset. During training, our model captures the correspondence between the set of 2D reference images and the 3D HOI motion. In the inference stage, it leverages the set of reference images synthesized by the text-to-image generation model to achieve precise motion control. Since the text-to-image generation model contains rich world knowledge, our model's generalization ability can be significantly enhanced.

\section{Experiment}

\begin{table*}[t!]
\centering
\renewcommand{\arraystretch}{0.95}
\setlength{\tabcolsep}{2.5pt}
\makebox[\textwidth][c]{
\resizebox{.95\textwidth}{!}{
\begin{tabular}{lcccccccccccc}
 \toprule 
 & \multicolumn{3}{c}{R-precision$\uparrow$} & \multicolumn{1}{c}{\multirow{2}{*}{FID$\downarrow$}} & \multicolumn{1}{c}{\multirow{2}{*}{Diversity $\uparrow$}} & \multicolumn{1}{c}{\multirow{2}{*}{FS$\downarrow$}} & \multicolumn{1}{c}{\multirow{2}{*}{$C_{prec}\uparrow$}} & \multicolumn{1}{c}{\multirow{2}{*}{$C_{rec}\uparrow$}} & \multicolumn{1}{c}{\multirow{2}{*}{$C_{F_1}\uparrow$}} & \multicolumn{1}{c}{\multirow{2}{*}{$C_{\%}$}} & \multicolumn{1}{c}{\multirow{2}{*}{$P_{hand}\downarrow$}} & \multicolumn{1}{c}{\multirow{2}{*}{MPJPE$\downarrow$}} \\
 \cmidrule(lr){2-4}
 \multirow{-2}{*}{Method} & Top-1 & Top-2 & Top-3 \\
 \midrule
 MDM~\cite{MDM}  & 0.31 & 0.53 & 0.66 & 1.56 & 11.15 & 0.36 & 0.69 & 0.51 & 0.54 & 0.46 & 0.57 & 23.11   \\
 ROG~\cite{ROG} & 0.33 & 0.54 & 0.68 & 1.54 & 11.44 & 0.41 & 0.69 & 0.57 & 0.57 & 0.51 & 0.57 & 22.59 \\
 CHOIS~\cite{CHOIS}   & 0.37 & 0.59 & 0.73 & 0.77 & 11.03 & 0.36 & 0.80 & 0.68 & 0.70 & 0.61 & 0.60 & 15.43   \\
 SemGeoMo~\cite{SemGeoMo}  & 0.36 & 0.62 & 0.76 & 0.79 & 10.35 & 0.36 & 0.82 & 0.74 & \textbf{0.75} & \textbf{0.66} & 0.59 & 16.34  \\
 \midrule
 MDM+ViHOI& 0.36 & 0.61 & 0.75 & 1.18& 10.97 & 0.32 & 0.67 & 0.55 & 0.57 & 0.50 & 0.57 & 23.12 \\
 ROG+ViHOI& 0.39 & 0.64 & 0.77 & \textbf{0.68} & \textbf{12.17} & 0.36 & 0.70 & 0.64 & 0.64 & 0.58 & \textbf{0.56} & 22.61 \\
 CHOIS+ViHOI& \textbf{0.41} & \textbf{0.65} & \textbf{0.79} & \textbf{0.68} & 11.57 & \textbf{0.26} & \textbf{0.83} & \textbf{0.75} & \textbf{0.75} &  0.64 & 0.58 & \textbf{14.97} \\
 \bottomrule
\end{tabular}
}
}

\caption{Quantitative comparisons on the FullBodyManipulation dataset~\cite{OMOMO}. We apply ViHOI as a plug-and-play module to three state-of-the-art HOI motion generation methods, demonstrating its effectiveness and flexibility.}
\vspace{-5mm}
\label{tab:result_on_OMOMO}
\end{table*}

\subsection{Datasets and Settings}

\paragraph{Datasets.}
We conduct experiments on two public 3D HOI motion datasets: FullBodyManipulation \cite{OMOMO} and BEHAVE \cite{BEHAVE}. The first dataset comprises 10 hours of HOI motion sequences. It covers 17 subjects and 15 objects. We follow the approach described in CHOIS \cite{CHOIS} to split the training and test data. Specifically, we use the data from 15 subjects as the training set and the remaining 2 subjects as the test set. The second dataset includes 1,451 HOI sequences, involving 8 subjects and 20 objects. We use the text annotations provided by HOI-Diff \cite{HOI-Diff} for each HOI sequence. We also strictly adhere to the official train-test split protocol \cite{BEHAVE}.

\vspace{-3mm}
\paragraph{Evaluation Metrics.}
Following existing works \cite{CHOIS,MDM}, we adopt the following evaluation metrics. For geometric accuracy, we measure pose fidelity using MPJPE (mean per-joint position error). It represents the mean Euclidean distance (in centimeters, cm) from the ground truth. To evaluate interaction quality, we assess the fidelity of the contact using Contact Precision (\textbf{$C_{prec}$}), Recall (\textbf{$C_{rec}$}), Contact Percentage (\textbf{$C_{\%}$}) and the F1 score (\textbf{$C_{F1}$}). For motion plausibility, we compute the Foot Sliding (FS) metric to quantify undesirable foot movement and the penetration score ($P_{hand}$) to measure mesh interpenetration, both as defined in \cite{CHOIS}. Moreover, we use the Fréchet Inception Distance (FID) to measure the distance between the generated and ground-truth motion distributions in the latent space. We also employ the R-score to evaluate the semantic alignment between the synthesized motion and the input text condition. Finally, we measure Diversity to assess the generational variety of our approach.

\begin{figure*}[t!]
  \centering
  \includegraphics[width=\linewidth,height=9.25cm]{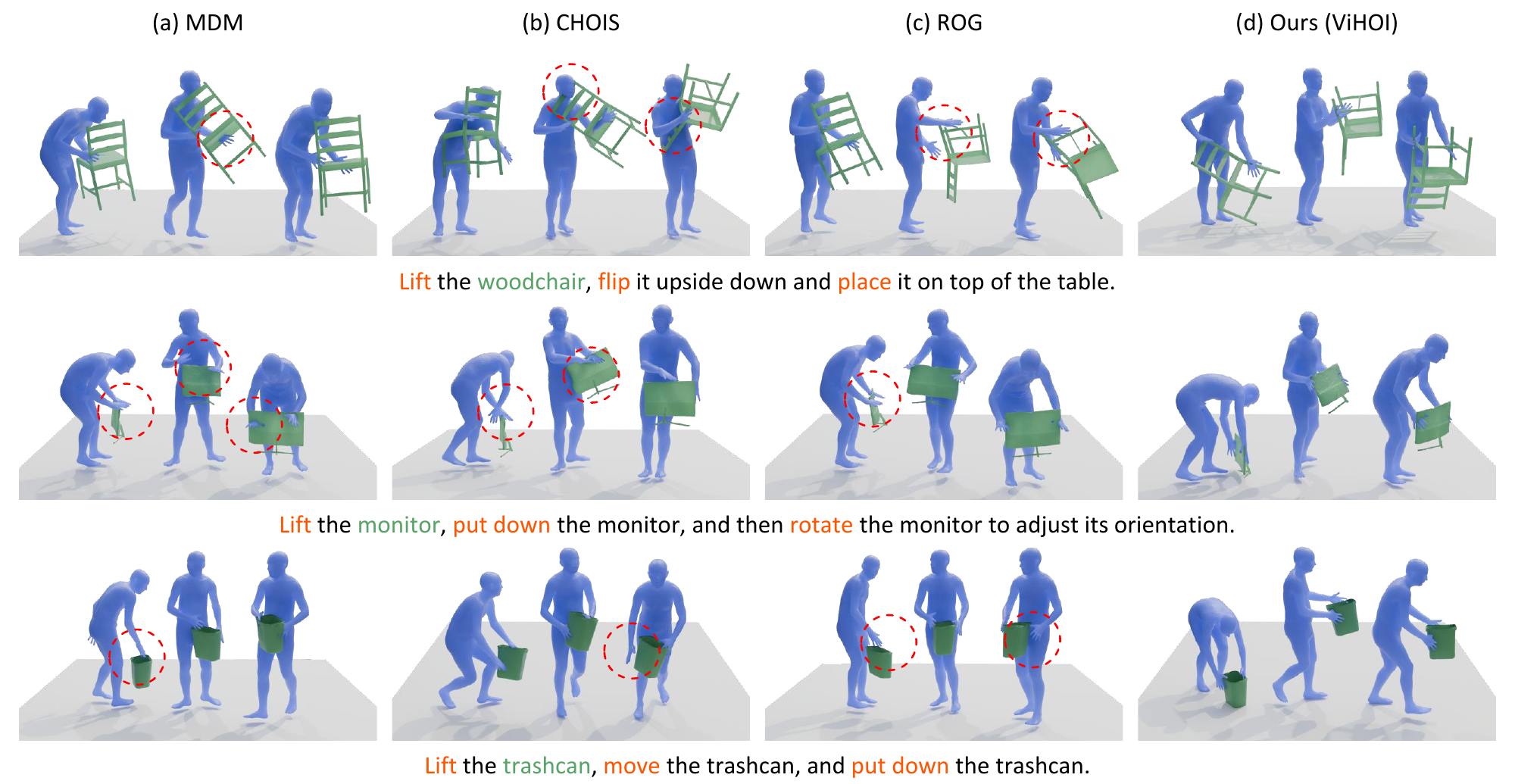}
  \vspace{-6mm}
  \caption{
  Qualitative comparisons on the FullBodyManipulation dataset~\cite{OMOMO}. Compared to state-of-the-art methods, our approach generates more realistic and physically plausible human-object interactions.
  }
  \label{fig:vis_baseline}
\end{figure*}

\vspace{-3mm}
\paragraph{Baselines.} We adopt the following state-of-the-art, open-source models as baselines: MDM \cite{MDM}, CHOIS \cite{CHOIS}, ROG \cite{ROG}, and SemGeoMo \cite{SemGeoMo}. We have two purposes. First, we showcase the versatility of our method as a plug-and-play module for these baselines. Second, we compare the performance of the different priors used in our work with that of the baselines. MDM is a widely used general text-to-motion generation model. We extend it by increasing its input and output dimensions to predict the object’s full 6D pose, thereby enabling HOI synthesis. CHOIS utilizes sparse object waypoints as a global path prior for HOI generation. ROG enriches the object’s geometric representation by sampling key points on its surface and jointly modeling the interactive distance field with human joints. SemGeoMo employs a dual prior enhancement strategy. It provides a semantic prior through fine-grained textual annotation generated by an LLM and a geometric prior via an affordance map and joint positions.

\begin{table*}[t!]
\centering
\renewcommand{\arraystretch}{0.95}
\setlength{\tabcolsep}{3.0pt}
\makebox[\textwidth][c]{
\resizebox{.98\textwidth}{!}{
\begin{tabular}{lcccccccccccc}
 \toprule 
 & \multicolumn{3}{c}{R-precision$\uparrow$} & \multicolumn{1}{c}{\multirow{2}{*}{FID$\downarrow$}} & \multicolumn{1}{c}{\multirow{2}{*}{Diversity$\uparrow$}} & \multicolumn{1}{c}{\multirow{2}{*}{FS$\downarrow$}} & \multicolumn{1}{c}{\multirow{2}{*}{$C_{prec}\uparrow$}} & \multicolumn{1}{c}{\multirow{2}{*}{$C_{rec}\uparrow$}} & \multicolumn{1}{c}{\multirow{2}{*}{$C_{F_1}\uparrow$}} & \multicolumn{1}{c}{\multirow{2}{*}{$C_{\%}$}} & \multicolumn{1}{c}{\multirow{2}{*}{$P_{hand}\downarrow$}} & \multicolumn{1}{c}{\multirow{2}{*}{MPJPE$\downarrow$}} \\
 \cmidrule(lr){2-4}
 \multirow{-2}{*}{Method} & Top-1 & Top-2 & Top-3 \\
 \midrule
MDM~\cite{MDM}  & 0.26 & 0.43 & 0.57 & 7.38 & \textbf{9.73} & 0.36 & 0.64 & 0.52 & 0.63 & 0.47 & 0.31 & 21.58   \\
ROG~\cite{ROG} & 0.18 & 0.38 & 0.54 & 8.08 & 9.22 &  0.33 & 0.68 & 0.48 & 0.60 & 0.48 & 0.35 & 19.71 \\
CHOIS~\cite{CHOIS}   & \textbf{0.35} & 0.51 & 0.68 & 4.99 & 9.11 & 0.40 & 0.83 & 0.71 & 0.74 & 0.65 & 0.27 & 15.42   \\
\midrule
CHOIS+ViHOI& 0.33 & \textbf{0.57} & \textbf{0.73} & \textbf{2.02} & 9.49 & \textbf{0.24} & \textbf{0.86} & \textbf{0.79} & \textbf{0.80} &  \textbf{0.71} & \textbf{0.19} & \textbf{14.58} \\
\bottomrule
\end{tabular}
}
} 
\vspace{-1mm}
\caption{Quantitative comparisons on the unseen objects of the FullBodyManipulation dataset~\cite{OMOMO}. We follow the data split strategy of \cite{Gen_HOI} and divide the dataset's objects by category, ensuring that the test set contains unseen objects.}
\vspace{-2mm}
\label{tab:result_on_UnseenObject_OMOMO}
\end{table*}

\begin{table}[!h]
\centering
\renewcommand{\arraystretch}{1.1}
\setlength{\tabcolsep}{1.0pt}
% 删除了冗余且容易导致排版错位的 \makebox[\textwidth][l]{ ... }
\resizebox{\columnwidth}{!}{ % 使用 \columnwidth 自动适应当前栏宽，比硬编码 .47\textwidth 更安全
\begin{tabular}{lccccccc} % 修正为实际的 8 列
 \toprule 
 & \multicolumn{3}{c}{R-precision $\uparrow$} & \multicolumn{1}{c}{\multirow{2}{*}{FID $\downarrow$}} & \multicolumn{1}{c}{\multirow{2}{*}{Diversity $\uparrow$}} & \multicolumn{1}{c}{\multirow{2}{*}{FS $\downarrow$}} &\multicolumn{1}{c}{\multirow{2}{*}{MPJPE $\downarrow$}} \\
 \cmidrule(lr){2-4}
 \multirow{-2}{*}{Method} & Top-1 & Top-2 & Top-3 \\
 \midrule
 MDM~\cite{MDM}  & 0.18 & 0.35 & 0.44 & 8.52 & \textbf{10.31} & 0.36  & 22.49   \\
 ROG~\cite{ROG} & 0.17 & 0.33 & 0.49 & 6.16 & 9.62 & 0.31  & 19.79  \\
 CHOIS~\cite{CHOIS} & \textbf{0.23} & 0.38 & 0.48 & 4.26 & 8.51 & 0.29  & 18.51   \\
 \midrule
 CHOIS+ViHOI& \textbf{0.23} & \textbf{0.41} & \textbf{0.51} & \textbf{3.90} & 9.94 & \textbf{0.26} & \textbf{17.09}  \\
 \bottomrule
\end{tabular}
} 
\vspace{-2mm}
\caption{Quantitative comparisons on the BEHAVE dataset~\cite{BEHAVE}.}
\vspace{-2mm}
\label{tab:result_on_BEHAVE}
\end{table}

\subsection{Quantitative and Qualitative Results}
\paragraph{Results on FullBodyManipulation.} Tab.~\ref{tab:result_on_OMOMO} presents our quantitative results on the FullBodyManipulation dataset~\cite{OMOMO}. We equip three baselines with our approach to explore visual and textual priors. ViHOI consistently improves the performance of baseline models across the vast majority of metrics. In particular, although ViHOI does not adopt the CLIP \cite{CLIP} text encoder, it still achieves high R-Precision in text matching. This demonstrates that our visual and textual priors carry rich semantic cues and strong semantic controllability. These results underscore ViHOI's versatility and ease of adaptation across existing HOI motion generation models.

SemGeoMo~\cite{SemGeoMo} introduces finer-grained text annotation and an affordance map as semantic and geometric priors. Although it performs well at modeling contacts, it still lags behind our method in FS and MPJPE. In contrast, our approach achieves superior results in both contact quality and joint accuracy, demonstrating that incorporating visual priors effectively balances local contact precision with global interaction realism.

In Fig.~\ref{fig:vis_baseline}, we present qualitative results by different models. Both ROG and MDM exhibit obvious object floating and drifting artifacts, which significantly degrade the realism of the interaction, while CHOIS suffers from severe penetration issues. In contrast, our model generates interactions that are more coherent with the textual descriptions and better aligned with real-world physical plausibility.

%Furthermore, to demonstrate the generalizability and flexibility of our approach, we employ the proposed VLM-based prior extractor as a plug-and-play module for diffusion-based HOI motion generators. Specifically, we integrate our framework into two representative models, MDM \cite{MDM} and ROG \cite{ROG}. The quantitative results are summarized in Tab.\ref{tab:result_on_OMOMO}, and visual comparisons are provided in Fig.xxx. Remarkably, both MDM and ROG show consistent improvements across most evaluation metrics, highlighting that our visual-language prior effectively enhances their generation quality without requiring architectural modifications. These results strongly underscore the versatility and ease of adaptation of our method across various HOI generation frameworks.
\vspace{-3mm}
\paragraph{Results on BEHAVE.} We further conduct experiments on the BEHAVE dataset~\cite{BEHAVE}. Since BEHAVE lacks ground-truth contact labels, we exclude contact‑related accuracy metrics from our evaluation metrics for this benchmark. As shown in Tab.~\ref{tab:result_on_BEHAVE}, ViHOI achieves strong performance on the BEHAVE dataset. Although MDM achieves higher scores on the diversity metric, its performance on other interaction-related metrics, such as FID and R-Precision, is noticeably inferior. We attribute this higher diversity score to its unstable generation behavior, in which the generated motions often fail to align with specific textual constraints. In contrast, our method strikes a better balance, producing interactions that are not only diverse but also consistently high-quality and semantically relevant.

\begin{table*}[h!]
\centering
\setlength{\tabcolsep}{3.0pt}
\makebox[\textwidth][c]{
\resizebox{0.95\textwidth}{!}{
\begin{tabular}{lcccccccccccc}
 \toprule 
 & \multicolumn{3}{c}{R-precision$\uparrow$} & \multicolumn{1}{c}{\multirow{2}{*}{FID$\downarrow$}} & \multicolumn{1}{c}{\multirow{2}{*}{Diversity$\uparrow$}} & \multicolumn{1}{c}{\multirow{2}{*}{FS$\downarrow$}} & \multicolumn{1}{c}{\multirow{2}{*}{$C_{prec}\uparrow$}} & \multicolumn{1}{c}{\multirow{2}{*}{$C_{rec}\uparrow$}} & \multicolumn{1}{c}{\multirow{2}{*}{$C_{F_1}\uparrow$}} & \multicolumn{1}{c}{\multirow{2}{*}{$C_{\%}$}} & \multicolumn{1}{c}{\multirow{2}{*}{$P_{hand}\downarrow$}} & \multicolumn{1}{c}{\multirow{2}{*}{MPJPE$\downarrow$}} \\
 \cmidrule(lr){2-4}
 \multirow{-2}{*}{Method} & Top-1 & Top-2 & Top-3 \\
\midrule
ViHOI-Pool& 0.15 & 0.24 & 0.32 & 26.03 & 7.63 & \textbf{0.23} & 0.51 & 0.37 & 0.40 & 0.35 & 0.68 & 22.62 \\
ViHOI-CLIP& 0.35 & 0.60 & 0.75 & 0.69& 10.33 & 0.30 & 0.80 & 0.69 & 0.71  & 0.60 & 0.60 &17.57 \\
\midrule
ViHOI& \textbf{0.41} & \textbf{0.65} & \textbf{0.79} & \textbf{0.68} & \textbf{11.57} & 0.26 & \textbf{0.83} & \textbf{0.75} & \textbf{0.75} & \textbf{0.64} & \textbf{0.58} & \textbf{14.97} \\
\bottomrule
\end{tabular}
}
}
\vspace{-2mm}
\caption{Ablation study on the FullBodyManipulation dataset \cite{OMOMO}. 
%ViHOI-Pool replaces the Q-Former-based adaptor using an average pooling layer, and ViHOI-CLIP encodes the textual instruction with the CLIP text encoder. 
We adopt CHOIS~\cite{CHOIS} as the HOI motion generator in this experiment.}
\vspace{-3mm}
\label{tab:ablation_study_1}
\end{table*}

\begin{table}[h!]
\centering
\setlength{\tabcolsep}{1.0pt}
% 删除了会导致双栏排版留白的 \makebox[\textwidth][l]
\resizebox{\columnwidth}{!}{ % 使用 \columnwidth 完美适配单/双栏的当前栏宽
\begin{tabular}{lcccccc} % 修正为实际的 7 列
 \toprule 
 Method & R-score$\uparrow$ & FID$\downarrow$ & $C_{prec}\uparrow$ & $C_{\%}$ & $P_{hand}\downarrow$ & MPJPE$\downarrow$ \\
 \midrule
V3-T24  & 0.67 & 2.35 & 0.81  & \textbf{0.67} & 0.56 & 18.05   \\
V3-T36 &  0.71 & 1.19 &  0.80 &  0.61 & 0.57 & 17.30 \\
V12-T12 & 0.75 & 0.87 &  0.81 &  0.62 & 0.59 & 15.90   \\
V12-T36 & 0.67 & 2.31 & 0.80 & 0.66 & 0.62 & 17.97   \\
V24-T24 & 0.61 & 3.15 &  0.81  & 0.64 & \textbf{0.54} & 16.94   \\
V36-T36  &  0.74 & 1.08 &  0.80  & 0.63 & 0.58 & 16.65  \\
T12-only &  0.72 & 1.28 &  0.81  & 0.62 & 0.59 & 17.49 \\
\midrule
Ours(V3-T12) & \textbf{0.79} & \textbf{0.68} & \textbf{0.83}   & 0.64 & 0.58 & \textbf{14.97} \\
\bottomrule
\end{tabular}
}
\vspace{-1mm}
\caption{Ablation study on the FullBodyManipulation dataset \cite{OMOMO}. We compare different layer combinations to extract $E_v$ and $E_t$. Vn-Tn indicates the LLM layer indices for the visual and textual modalities, respectively.}
\vspace{-2mm}
\label{tab:ablation_study_2}
\end{table}

\vspace{-3mm}
\paragraph{Results on Unseen Objects.} Following the data split strategy of \cite{Gen_HOI}, we divide the FullBodyManipulation dataset by object category, ensuring that the test set contains unseen objects. This setup enables a fair evaluation of generalization to novel objects. As shown in Tab.~\ref{tab:result_on_UnseenObject_OMOMO}, ViHOI exhibits no significant performance degradation when generating interactions with unseen objects. Moreover, it consistently and significantly outperforms other models in this setting. This result demonstrates the robust generalization capability of our method. 

\begin{figure}[ht!]
  \centering
  \includegraphics[width=\linewidth]{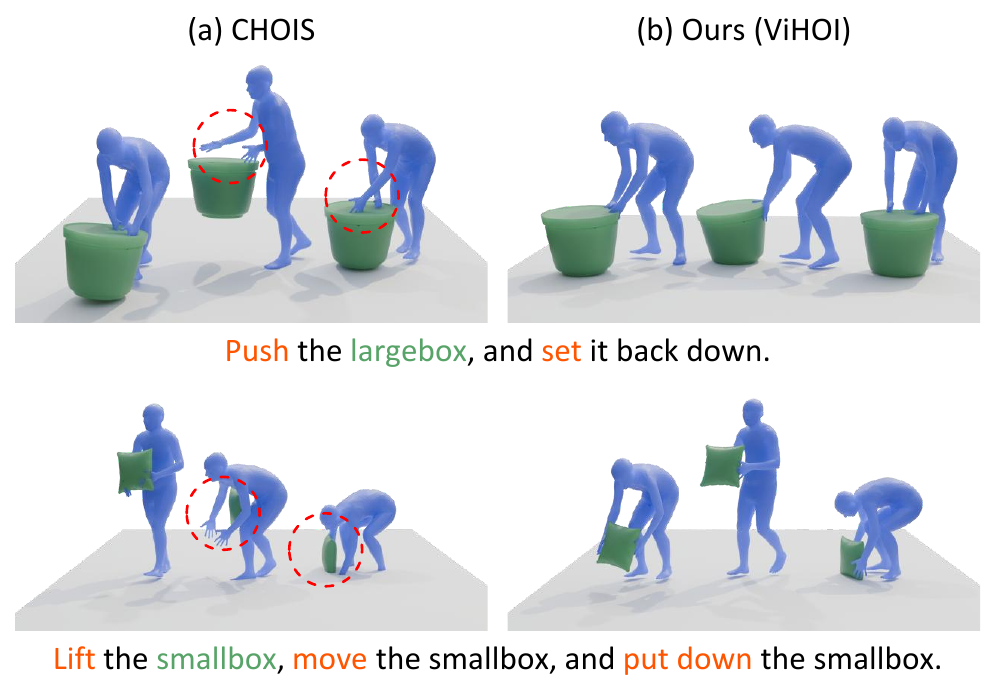}
  \vspace{-5mm}
  \caption{ Qualitative comparisons on the 3D-Future dataset \cite{3D-Future}. Our approach generates more realistic human-object interactions on unseen objects.
  }
  \vspace{-2mm}
  \label{fig:unseen_object}
\end{figure}

To further assess ViHOI's generalization ability, we also conduct experiments on the 3D‑FUTURE dataset \cite{3D-Future}. Since this dataset contains only object meshes, we follow the approach of \cite{CHOIS} and replace the objects in the ground‑truth (GT) sequences with 3D‑Furniture. As illustrated in Fig.~\ref{fig:unseen_object}, our method continues to produce plausible and coherent motions. Experiments on both datasets confirm that leveraging reference images generated by a text-to-image model enhances the generalization of HOI motion generation models.

\subsection{Ablation Studies}
To assess the contribution of each component within ViHOI, we conduct a series of ablation studies to evaluate their individual impacts. We systematically dissect our framework to isolate the contributions of our two core innovations: (1) leveraging the Q-Former to compress the dense, high-dimensional VLM priors, and (2) extracting textual priors from the VLM. For the first setting, we apply average pooling directly to the VLM's visual and textual embeddings, respectively, and then project the resulting vectors through a simple MLP to align with the feature dimension of the motion generator. For the second setting, we adopt CLIP \cite{CLIP} as the text encoder, and combine the text tokens produced by CLIP with the visual priors extracted from the VLM as conditioning inputs. 

As shown in Tab.~\ref{tab:ablation_study_1}, when the VLM embeddings are directly averaged through simple pooling, the overall model performance drops sharply. 
%Although it achieves a relatively low score on FS, the overall performance degrades significantly. 
This pronounced decline underscores the necessity of the Q‑Former-based prior adaptor, which effectively extracts and structures interaction cues from the VLM’s intermediate representations. Compared with CLIP-based text encoding, our method achieves superior performance across both text-matching accuracy and interaction quality. This demonstrates the effectiveness of extracting textual instructions directly from the LLM component of a VLM, enabling richer semantic understanding and more precise motion control.

Moreover, we compare different LLM layer combinations to extract $E_v$ and $E_t$ from VLM as conditioning signals. As shown in Tab.~\ref{tab:ablation_study_2}, the combination of the 3rd layer for visual and the 12th layer for textual yields the best performance. Moreover, it significantly outperforms the variant that uses the textual priors alone, i.e., `T12-only' in Tab.~\ref{tab:ablation_study_2}. These results highlight the effectiveness of our decoupled prior-extraction strategy for leveraging visual and textual conditions.

\section{Conclusion and Limitations} 
In this work, we propose ViHOI, a novel plug‑and‑play framework that enables HOI motion diffusion models to leverage rich visual priors from 2D images effectively.
%, achieving high‑quality synthesis of HOI sequences. 
Our approach employs a large VLM as a prior‑extraction engine and adopts a layer-decoupled strategy to obtain complementary visual and textual priors. Q-Former-based prior adaptors then distill these high-dimensional representations into compact tokens, providing semantically consistent conditioning signals for the diffusion model. Trained on motion‑rendered images from the dataset, ViHOI utilizes reference images synthesized by a text‑to‑image generation model during inference, enabling strong generalization to unseen objects and delivering superior results across multiple benchmarks. 
One limitation of our current work is the lack of fine‑grained hand annotations in the datasets used by our method, which prevents the model from accurately generating detailed finger-motion sequences.

\vspace{-3mm}
\paragraph{Broader Impacts.} The synthesis of realistic and controlled HOI sequences holds significant potential to advance applications in virtual reality, robotics, and digital entertainment. To the best of our current knowledge, this research does not present any obvious negative social impacts.

\vspace{-3mm}
\paragraph{Acknowledgement.} This work was supported by the National Natural Science Foundation of China under Grant 62476099 and 62076101, Guangdong Basic and Applied Basic Research Foundation under Grant 2024B1515020082 and 2023A1515010007, the Guangdong Provincial Key Laboratory of Human Digital Twin under Grant 2022B1212010004, the TCL Young Scholars Program.
\vspace{-8mm}
{   
    \small
    \bibliographystyle{ieeenat_fullname}
    \bibliography{main}
}
\clearpage
\setcounter{page}{1}
\maketitlesupplementary

\setcounter{section}{0}
\renewcommand{\thesection}{\Alph{section}}

\setcounter{subsection}{0}
\renewcommand{\thesubsection}{\thesection.\arabic{subsection}}

\setcounter{figure}{0}
\setcounter{table}{0}
\setcounter{equation}{0}
\renewcommand{\thefigure}{\arabic{figure}}
\renewcommand{\thetable}{\arabic{table}}
\renewcommand{\theequation}{\arabic{equation}}

\section{Overview}
\label{sec:Overview}
This supplementary material provides a comprehensive description of our approach, including method details (Section~\ref{sec:meth-detail}), evaluation details (Section~\ref{sec:eval-detail}), additional experiments (Section~\ref{sec:addi-exper}), additional visualization results (Section~\ref{sec:addi-vis-resu}), and analysis of VLM understanding (Section~\ref{sec:anal-vlm}).

\section{Method Details}
\label{sec:meth-detail}
In our framework, we adopt the same object geometry representation as used in the downstream generator. Taking the framework diagram in the main text as an example, we choose CHOIS~\cite{CHOIS} as our HOI generator, and the geometric shape of the object is encoded using Basis Point Set (BPS)~\cite{BPS} and then projected into a 256‑dimensional embedding space through an MLP. This 256‑dimensional geometric embedding is fused with the motion tokens and subsequently concatenated along the temporal dimension with both the visual and textual prior tokens. The transformer's self‑attention layers jointly process the combined sequence.

In our comparison experiments, we follow the original object-geometry handling of each downstream HOI generator. Specifically, when integrating our framework with MDM~\cite{MDM} and ROG~\cite{ROG}, we adopt the exact geometry representation used in their original implementations. The MDM version we use is the one released by the ROG authors, where both models represent the object using 24 surface keypoints. These keypoints are obtained by combining two sampling strategies on the object mesh surface: 8 boundary keypoints aligned with the object's Axis-Aligned Bounding Box~\cite{AABB} that capture its global extent, and 16 keypoints obtained via Poisson Disk Sampling~\cite{PDS} that preserve finer geometric details. This ensures that any observed performance improvements are attributable solely to our proposed priors, rather than discrepancies in geometric representation.

Regarding the training details, we strictly follow the original training recipes of the respective baselines (e.g., MDM and CHOIS). The only architectural modification is the replacement of their original CLIP text encoder with our proposed VLM and Q-Former module. To handle the scale discrepancies and distribution shifts across the end-layer features of different methods, the Q-Former adapter is jointly trained with each specific baseline generator. Because of this joint training strategy, the visual and textual prior tokens $C_v$ and $C_t$ are dynamically optimized. This ensures that the generated tokens seamlessly align with the specific feature distributions of each generator.

\section{Evaluation Details}
\label{sec:eval-detail}
Currently, most feature extractors used to evaluate Human-Object Interaction (HOI) motions primarily focus on human poses, neglecting the spatial positions and rotational dynamics of the involved objects. To overcome this limitation, we draw inspiration from the T2M~\cite{T2M} framework and adopt a similar evaluation protocol. In our approach, a frozen CLIP text encoder~\cite{CLIP} is employed to transform textual descriptions into feature embeddings. Meanwhile, the generated HOI motion sequences are processed using a bidirectional GRU (BiGRU) model. To ensure that the evaluation metrics accurately capture the quality of the generated motions, we adjust the BiGRU model's input dimensionality to meet the parameter requirements for HOI sequence visualization. Specifically, we set the input dimension to 147: the first 3 dimensions represent the root joint parameters of the human body, 132 dimensions correspond to the 6D relative rotations of 22 joints, 3 dimensions encode the object’s translation parameters, and the remaining 9 dimensions describe the object’s rotation matrix. By minimizing the feature distance between matched text–HOI pairs, our method effectively builds a robust alignment between natural language descriptions and HOI motion sequences.

\section{Additional Experiments}
\label{sec:addi-exper}
\paragraph{Impact of Query Quantity on Prior Adaptor.}

Our previous Prior Adaptor module extracts interaction priors from a large vision–language model (VLM) using learnable queries. To analyze the impact of query quantity, we evaluate different numbers of queries. As shown in Table~\ref{tab:impact_of_query}, employing a single query achieves the best performance across most metrics. This observation aligns with our design philosophy: the VLM-based prior is intended to capture a compact, global cue of human–object interaction. Introducing more queries expands the latent prior’s dimension. It forces the model to attend to multiple prior tokens simultaneously, potentially diluting the semantic signal and introducing redundant or less informative visual features. Therefore, we adopt a single-query configuration that provides stable, semantically coherent prior cues.

\begin{table}[t]
\centering
\setlength{\tabcolsep}{1.5pt}
% 去掉 \makebox，使用 \columnwidth 自动适配栏宽
\resizebox{\columnwidth}{!}{ 
\begin{tabular}{lcccccc} % 修正为实际的 7 列
 \toprule 
 Method & R-score$\uparrow$ & FID$\downarrow$ & $C_{prec}\uparrow$ & $C_{\%}$ & $P_{hand}\downarrow$ & MPJPE$\downarrow$ \\
 \midrule
 k = 2  & 0.72 & 1.30 & 0.71  & 0.61 & 0.60 & 17.52   \\
 k = 4 &  0.63 & 3.07 & 0.78  &  0.58 & 0.59 & 18.38 \\
 k = 8 & 0.69 & 2.81 & 0.77 &  0.57 & 0.61 & 17.94   \\
 \midrule
 k = 1 & \textbf{0.79} & \textbf{0.68} & \textbf{0.83} & \textbf{0.64} & \textbf{0.58} & \textbf{14.97} \\
 \bottomrule
\end{tabular}
}
\caption{Impact of Adaptor Query Number (k) on Generation Performance. We vary only the number of visual prior queries, while keeping the number of text queries fixed at one.}
% 修复了拼写错误并去掉了空格
\label{tab:impact_of_query}
\end{table}

\begin{figure*}[t!]
  \centering
  \includegraphics[width=\linewidth]{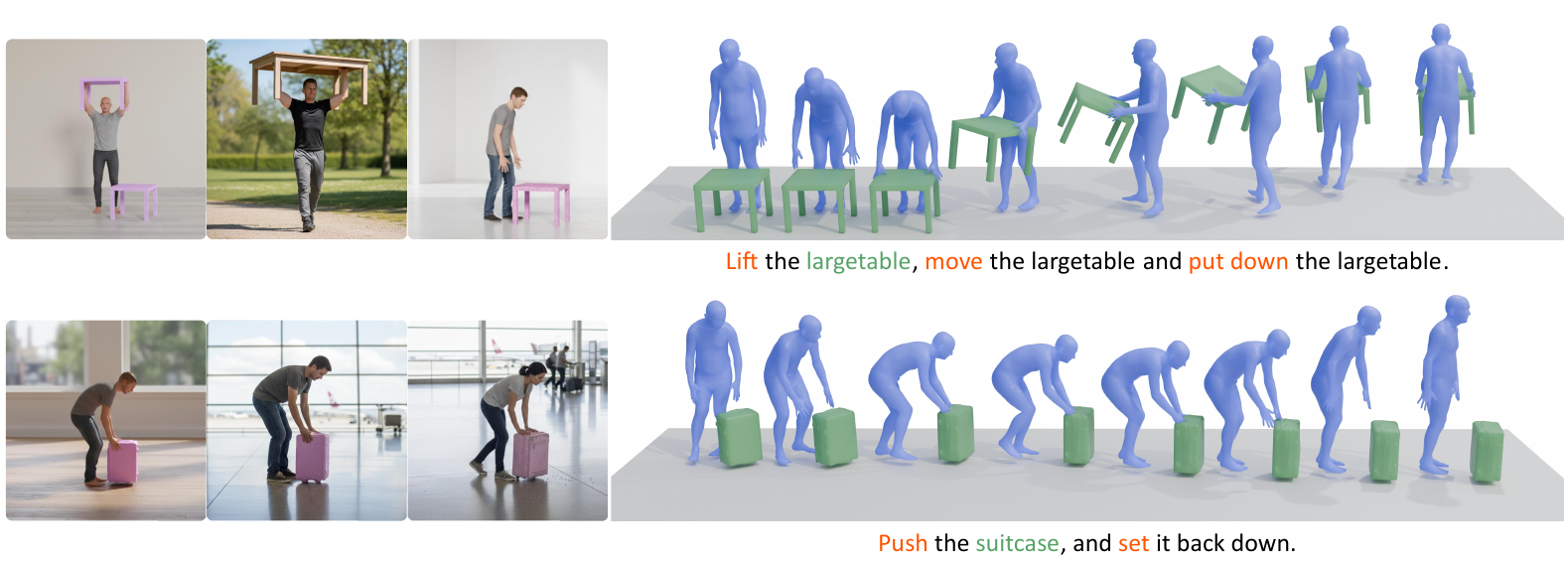}
  \caption{Qualitative results on the FullBodyManipulation dataset~\cite{OMOMO}. The three images on the left side are the reference inputs, while the right side shows the motion sequences generated from them. Despite imperfections in these reference images, the generated HOI motions remain plausible and well aligned with the textual semantics.}
  \label{fig:supp_rf}
\end{figure*}

\begin{table}[t]
\centering
\setlength{\tabcolsep}{1.5pt}
% 同样去掉了多余的 \makebox，改用更安全的 \columnwidth
\resizebox{\columnwidth}{!}{ 
\begin{tabular}{lcccccc} % 修正为实际的 7 列
 \toprule 
 Method & R-score$\uparrow$ & FID$\downarrow$ & $C_{prec}\uparrow$ & $C_{\%}\uparrow$ & $P_{hand}\downarrow$ & MPJPE$\downarrow$ \\
 \midrule
 ViHOI-GT & \textbf{0.79} & \textbf{0.29} & 0.82 &  \textbf{0.64} & 0.59 & \textbf{12.94}   \\
 ViHOI & \textbf{0.79} & 0.68 & \textbf{0.83} & \textbf{0.64} & \textbf{0.58} & 14.97 \\
 \bottomrule % 将原来的 \midrule 修正为了 \bottomrule
\end{tabular}
}
\caption{Impact of reference images. ViHOI-GT uses images rendered from GT motion, while ViHOI uses images from the T2I generation model.}
% 修复了拼写错误并去掉了空格，引用时请用 \ref{tab:impact_of_reference_images}
\label{tab:impact_of_reference_images} 
\end{table}

\paragraph{Impact of Text-to-Image Generation Model.}
During inference, our method leverages reference images generated by a text-to-image generation model~\cite{Nano_banana} to provide visual prior information. Although these reference images may occasionally exhibit appearance flaws or unrealistic renderings, we observe that such visual imperfections have little impact on the final quality of HOI generation. We believe this is because our Prior Adaptor emphasizes capturing high-level semantic relationships within the image rather than low-level pixel details. Moreover, the textual priors extracted from the prompts offer a holistic description of the intended action, ensuring that the generated human–object interactions remain globally coherent and semantically accurate.

To assess ViHOI's robustness to the quality of reference images, we compared its performance under two distinct settings: (1) using reference images rendered from the Ground-Truth (GT) motion from the test set, and (2) using images generated by the Text-to-Image (T2I) generation model described in Section 3.3. As shown in Tab.~\ref{tab:impact_of_reference_images}, using T2I-generated images results in a certain degree of performance drop compared to those rendered from ground-truth (GT) motions. Still, the decline remains within a reasonable and acceptable range. Crucially, even in this more challenging setting, our method still outperforms existing state-of-the-art models, demonstrating its strong adaptability to variations in image style and rendering quality. To maintain a strict evaluation protocol and prevent test data leakage, we exclusively use the T2I generation model to produce reference images in all our main comparative experiments.

Fig.~\ref{fig:supp_rf} further provides qualitative evidence of this robustness. Even when the reference images contain noticeable artifacts or implausible visual effects, our model successfully produces accurate interaction trajectories and physically plausible contact patterns between humans and objects. These qualitative results demonstrate that our approach is highly resilient to variations in image quality, relying primarily on semantic cues rather than photorealistic fidelity.

\begin{figure}[t]
  \centering
  \includegraphics[width=1.0\linewidth]{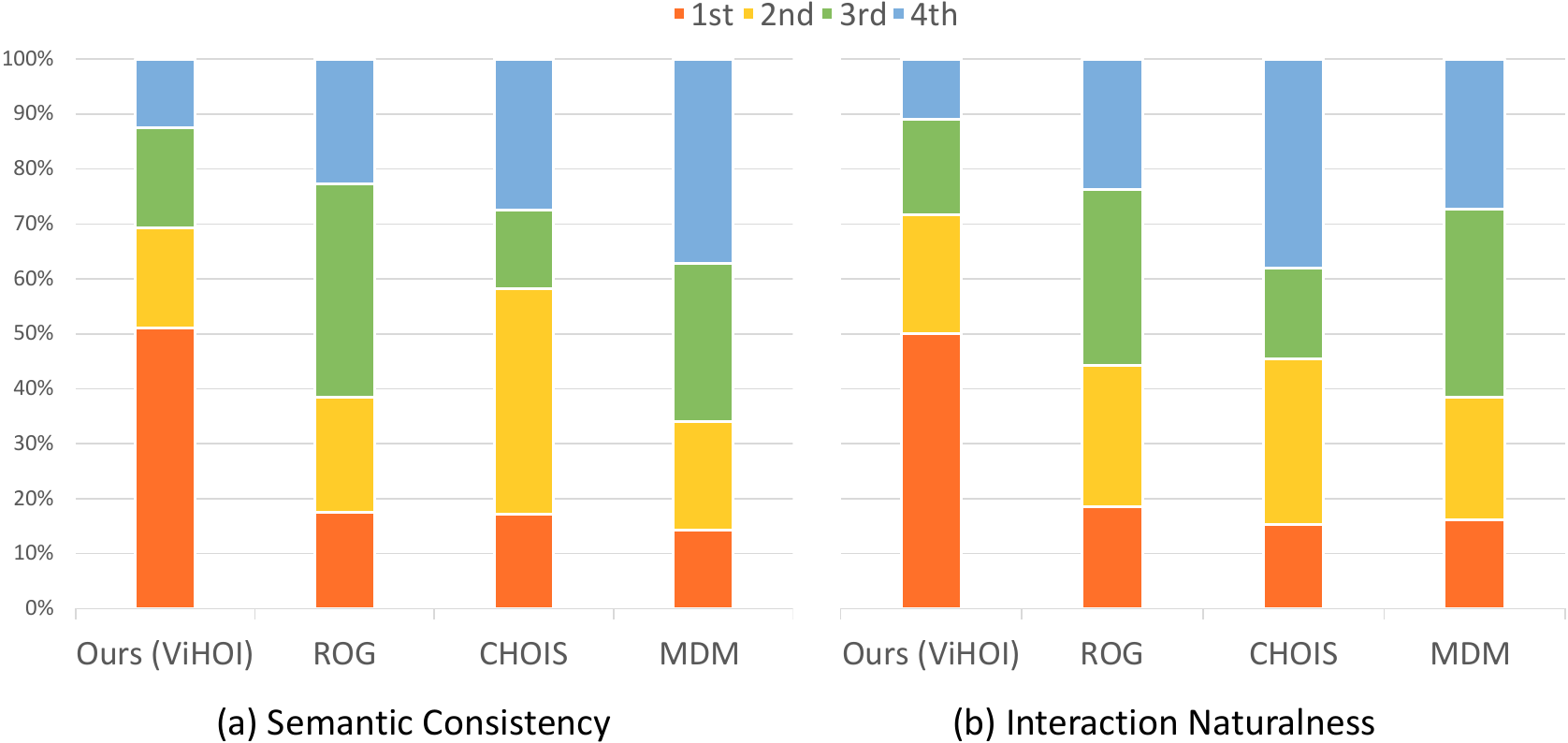}
  \caption{User study on the FullBodyManipulation dataset~\cite{OMOMO}.
  }
   \vspace{-2mm}
  \label{fig:us-nature}
\end{figure}

\begin{figure}[t]
  \centering
  \includegraphics[width=0.55\linewidth]{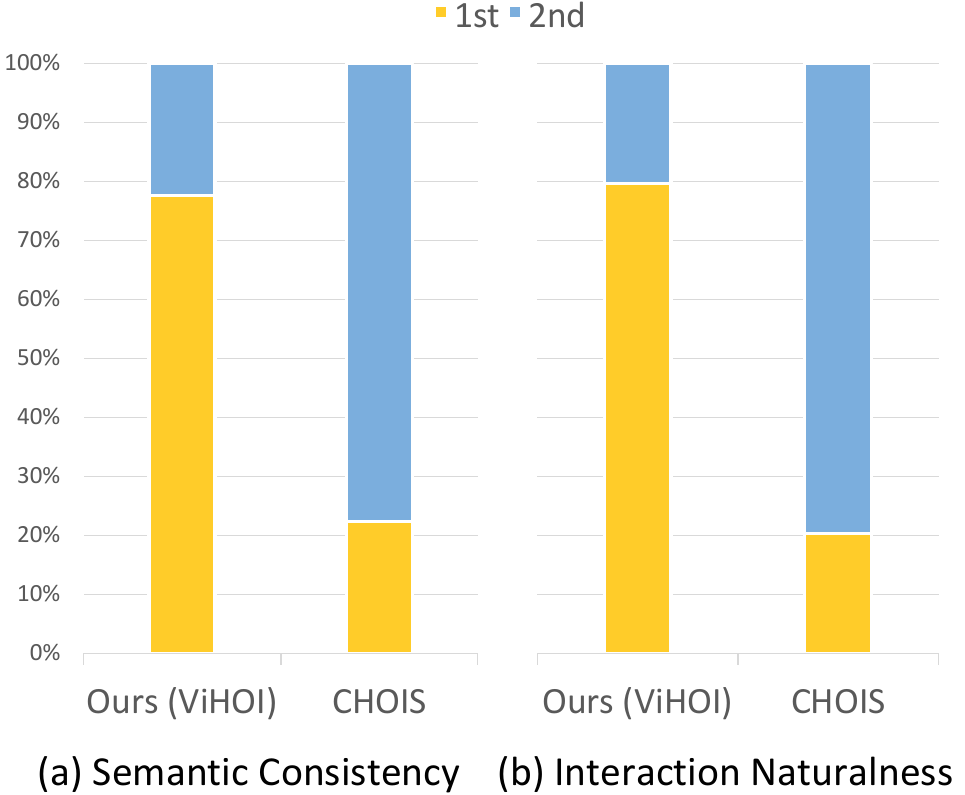}
  \caption{User study on the 3D-Future dataset~\cite{3D-Future}.
  }
   \vspace{0mm}
  \label{fig:us-unseen}
\end{figure}

\begin{figure*}[t!]
  \centering
  \includegraphics[width=\linewidth]{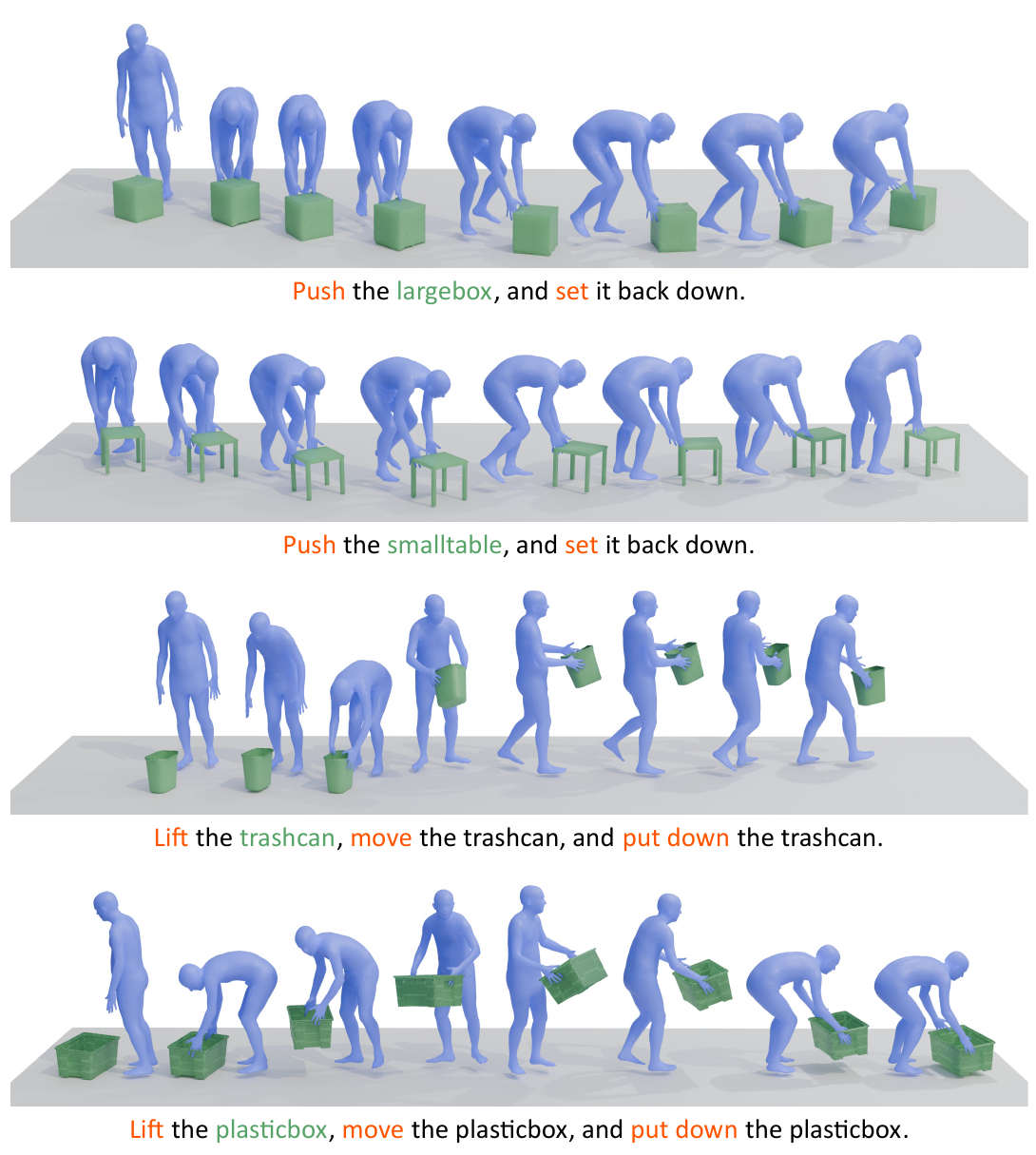}
  \caption{Additional visualization result on the FullBodyManipulation dataset~\cite{OMOMO}.}
  \label{fig:supp_nature}
\end{figure*}

\vspace{-5mm}
\paragraph{Impact of VLM and T2I on Computational Overhead.}
We evaluate the computational overhead of our framework using a single RTX 3090 GPU. The VLM introduces a marginal overhead of only 0.65s, whereas the Text-to-Image API call requires 7.20s. Although these modules introduce a certain degree of inference latency, they yield significant performance improvements. Furthermore, due to the stochastic nature of diffusion models, we emphasize that for the same textual instruction, a single generated reference image can be reused to generate multiple diverse HOI motions without repeating the time-consuming T2I process.

\paragraph{User Study.}
To evaluate the perceptual quality of our method, we conduct a user study comparing ViHOI against three baseline methods~\cite{MDM, ROG, CHOIS} on 20 text prompts from the FullBodyManipulation dataset~\cite{OMOMO}. Furthermore, to specifically assess generalization capabilities, we compare ViHOI against CHOIS~\cite{CHOIS} using 10 unseen objects from the 3D-Future dataset~\cite{3D-Future}.

\begin{figure*}[t!]
  \centering
  \includegraphics[width=\linewidth]{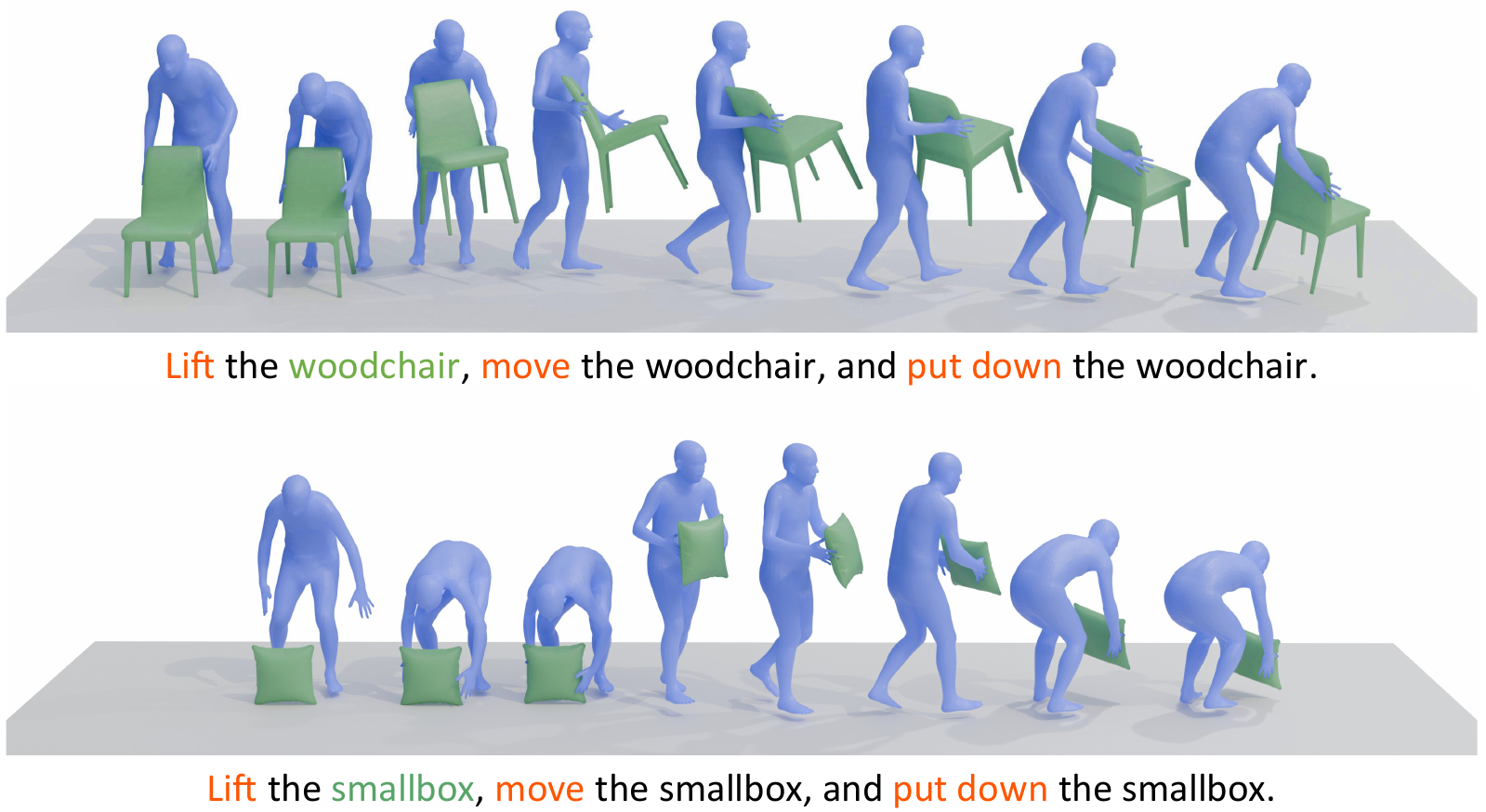}
  \caption{Additional visualization result on the 3D-FUTURE dataset~\cite{3D-Future}.}
  \label{fig:supp_unseen}
\end{figure*}

Following the evaluation protocol established by~\cite{ROG}, we asked a total of 20 participants to rank the results according to two criteria: (1) Semantic Consistency (alignment between animations and text descriptions), and (2) Interaction Naturalness (naturalness of poses and object interactions). As illustrated in Fig.~\ref{fig:us-nature} and~\ref{fig:us-unseen}, our method received significantly higher ratings in both generation quality and generalization. Participants consistently favored ViHOI for its superior text-motion alignment, more plausible interactions, and remarkable ability to generalize to unseen objects.

\section{Additional Visualization Results.}
\label{sec:addi-vis-resu}
To demonstrate the diversity and effectiveness of our approach, we present additional HOI generation results across various scenarios.

\paragraph{More Generation Results.}
We present additional generation results in Fig.~\ref{fig:supp_nature}, demonstrating the quality of our method across different actions and object categories, and include a detailed comparison and demonstration with three baseline methods~\cite{MDM, ROG, CHOIS} on the FullBodyManipulation dataset~\cite{OMOMO} in our accompanying video. These examples highlight ViHOI’s ability to generate natural and semantically accurate interactions across different actions and object categories.

\paragraph{More Results on Unseen Objects.}
We present additional results on unseen objects from the 3D-FUTURE dataset~\cite{3D-Future} in Fig.~\ref{fig:supp_unseen} and include detailed comparison and demonstration with CHOIS~\cite{CHOIS} in the accompanying video. These examples show that ViHOI can maintain natural and accurate interaction generation even when encountering previously unseen objects.

\section{Analysis of VLM Understanding}
\label{sec:anal-vlm} 
\begin{figure*}[h!]
  \centering
  \includegraphics[width=\linewidth]{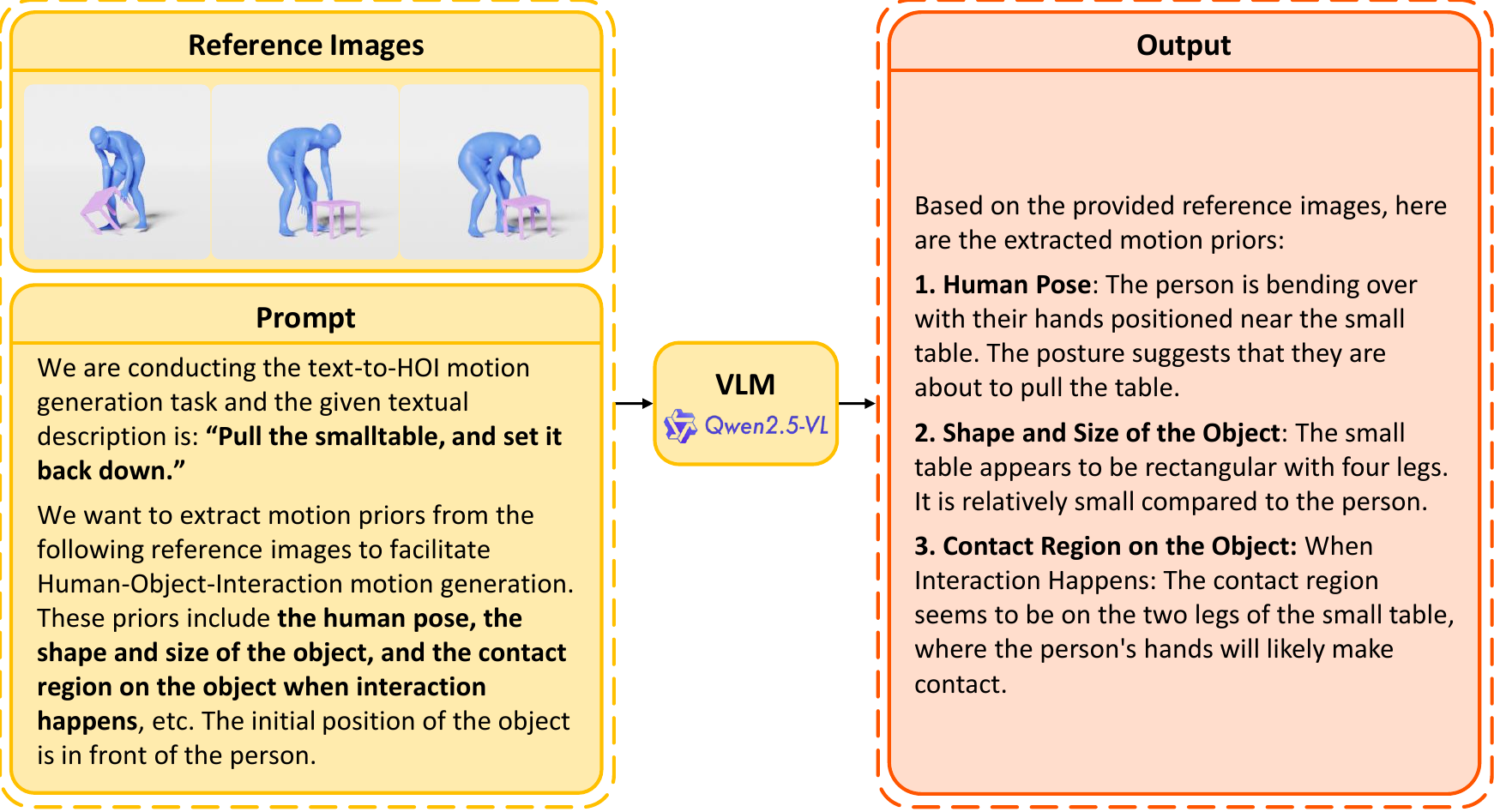}
  \caption{Qualitative analysis of VLM understanding. The text annotation is ``Pull the smalltable, and set it back down."}
 \vspace{13mm}
    \label{fig:VLM_pipeline_1}
\end{figure*}

\begin{figure*}[h!]
  \centering
  \includegraphics[width=\linewidth]{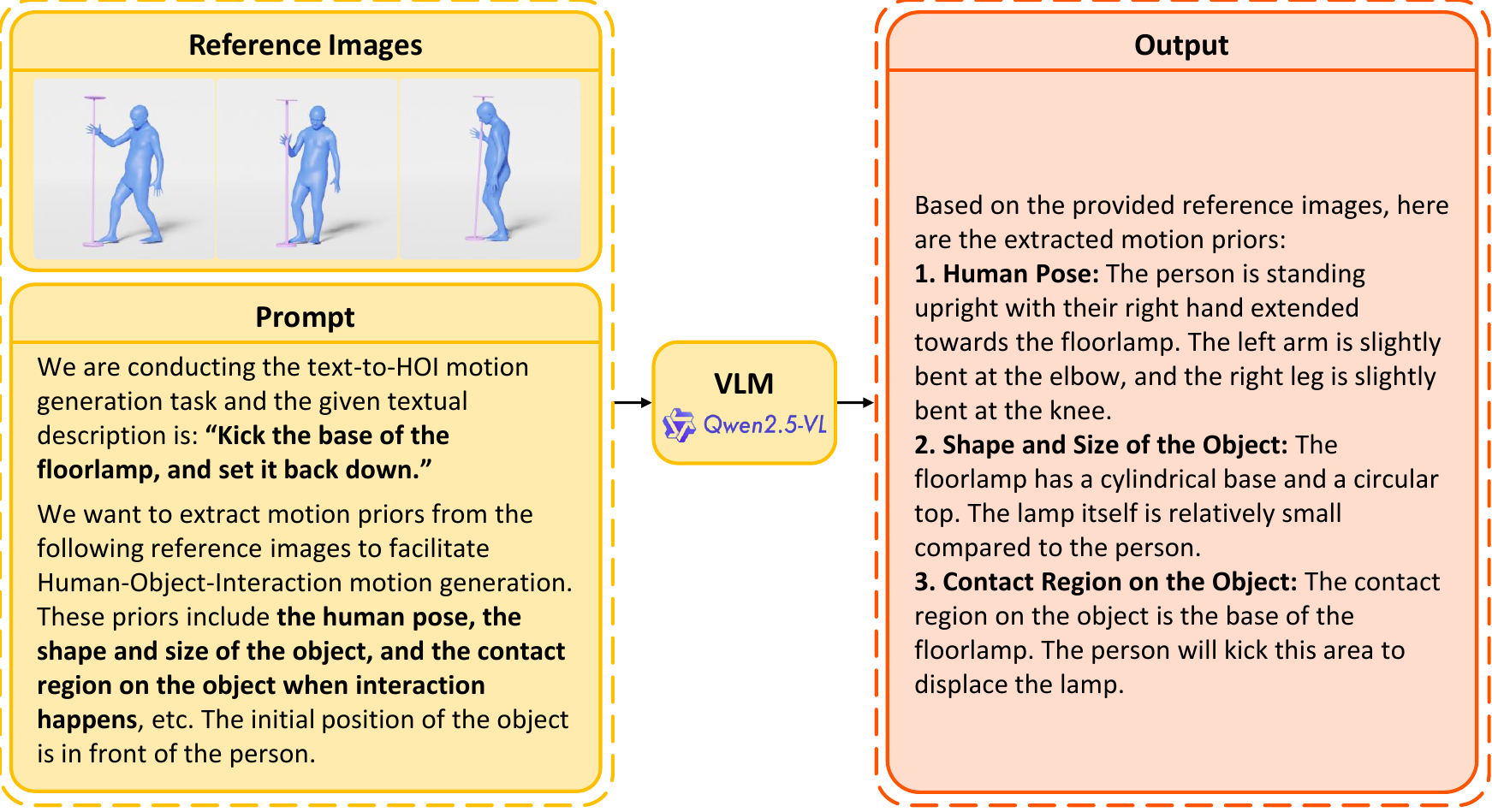}
  \caption{Qualitative analysis of VLM understanding. The text annotation is ``Kick the base of the floorlamp, and set it back down."}
  \label{fig:VLM_pipeline_2}
\end{figure*}

To better illustrate the semantic capacity of the VLM we use for prior extraction, we present several examples of its autoregressive textual outputs given our rendered reference images and prompts. Although our method only uses intermediate-layer embeddings rather than the final decoded text, these outputs demonstrate that the VLM reliably captures high-level human–object relations. This supports the rationale behind using VLM embeddings as interaction priors in our model.
As illustrated in Fig.~\ref{fig:VLM_pipeline_1} to~\ref{fig:VLM_pipeline_2} , the VLM is able to infer meaningful priors—such as the human's standing posture, the height and structure of the floorlamp, and the likely contact region at the flooorlamp’s base—directly from the reference images. These decoded outputs demonstrate that the VLM preserves high-level relational semantics, reinforcing the effectiveness of using its intermediate representations as motion priors.

% --- 附录部分 ---
% \input{sec/X_suppl}
% {
%     \small
%     \bibliographystyle{ieeenat_fullname}
%     \bibliography{main}
% }

% % WARNING: do not forget to delete the supplementary pages from your submission 

\end{document}